\documentclass[11pt]{article}

\usepackage{agenticlearning}
\usepackage{hyperref}
\usepackage{url}
\usepackage{graphicx}
\usepackage{bbold}
\usepackage{bbm}
\usepackage{booktabs}
\usepackage{multirow}
\usepackage{wrapfig}
\usepackage{titlesec}
\let\subparagraph\llncssubparagraph
\usepackage{bm}
\usepackage{subcaption,siunitx,booktabs}
\usepackage{enumitem}
\usepackage{multicol}

\usepackage{mathtools}
\usepackage{titlesec}
\usepackage{setspace}
\usepackage[most]{tcolorbox}
\usepackage{xcolor}

\usepackage{microtype}
\usepackage{hyperref}
\usepackage{url}
\usepackage{booktabs}

\usepackage{amsmath}
\usepackage{amssymb}
\usepackage{mathtools}
\usepackage{amsthm}
\usepackage{bm}
\usepackage{subcaption,siunitx,booktabs}
\usepackage{enumitem}
\usepackage{multicol}
\usepackage{hyperref}
\usepackage{url}
\usepackage{graphicx}
\usepackage{bbold}
\usepackage{bbm}
\usepackage{booktabs}
\usepackage{multirow}
\usepackage{wrapfig}
\usepackage{titlesec}

\definecolor{softbeige}{RGB}{250,245,230}
\definecolor{softbeigedark}{RGB}{230,220,200}

\newtcolorbox{takeaway}[1][]{
  colback=softbeige,
  colframe=softbeigedark,
  boxrule=0.4pt,
  arc=3pt,
  left=2pt, right=2pt, top=2pt, bottom=2pt,
  before skip=8pt, after skip=8pt,
  fonttitle=\small\bfseries\color{black},
  coltitle=black,
  colbacktitle=softbeigedark,
  title=#1,
  title style={top=4pt, bottom=4pt},
}

\definecolor{QwenThreeBase}{HTML}{87CEEB}
\definecolor{QwenThree}{HTML}{1f77b4}
\definecolor{QwenTwoFiveBase}{HTML}{FFB347}
\definecolor{QwenTwoFive}{HTML}{d62728}
\definecolor{LlamaThreeBase}{HTML}{A5D6A7}
\definecolor{LlamaThree}{HTML}{2ca02c}
\definecolor{DeepSeek}{HTML}{9467bd}

\begin{document}

\shorttitle{When Does Verification Pay Off? A Closer Look at LLMs as Solution Verifiers}
\shortauthor{Lu \etal}

% \makeatletter
% \def\blfootnote{\gdef\@thefnmark{}\@footnotetext}
% \makeatother

\makeatletter
\newcommand{\blfootnote}[1]{%
  \begingroup
  \renewcommand\@makefntext[1]{##1}% Remove indentation + marker formatting
  \footnotetext{#1}%
  \endgroup
}
\makeatother

\title{When Does Verification Pay Off?\\
A Closer Look at LLMs as Solution Verifiers}
\author{
  Jack Lu$^{*}$, Ryan Teehan$^{*}$, Jinran Jin, and Mengye Ren \\
  Agentic Learning AI Lab, New York University \\
  \texttt{\{yl11330, rst306, jj3007, mengye\}@nyu.edu}\\
  \url{https://agenticlearning.ai/llm-verification}
}
\date{November 20, 2025
\blfootnote{* Equal contribution.}
}
\maketitle

\begin{abstract}
Large language models (LLMs) can act as both problem solvers and solution verifiers, where the latter select high-quality answers from a pool of solver-generated candidates. This raises the question of under what conditions verification pays off in solver–verifier systems. Prior work has conducted only limited studies of the factors influencing verification performance, focusing primarily on self-verification and examining neither the relationship between solver and verifier model families nor the effects of reasoning post-training. To rectify this, we present a systematic study across 37 models spanning multiple families, sizes, and base vs. post-trained variants, evaluated on 9 benchmarks covering logical reasoning, structured puzzles, symbolic computation, mathematics, commonsense, factual recall, and domain knowledge. In order to support our analysis, we introduce and empirically validate \textit{verifier gain}, a metric that predicts the performance improvements from \textit{test-time verifier-based rejection sampling}. Our experiments find that 1) verification across model families is more effective than either self-verification or verification within the same family, and more generally that the benefits of verification decrease as the solver and verifier become more similar, 2) reasoning post-training weakens self-improvement abilities but strengthens cross-family improvement, and 3) some tasks are inherently more amenable to improvement through verification, particularly mathematical and logical tasks. 
\end{abstract}

\section{Introduction}
\looseness=-10000
Problem-solving with LLMs has progressed beyond querying a standalone model for a solution to a system where generated solutions are verified by other models. Test-time solution verification is integral to a variety of concrete approaches, including simple strategies, such as filtering multiple generated candidates~\citep{zhao2025sample}, as well as more complex strategies, such as iterative refinement~\citep{madaan2023selfrefine}. With test-time solution verification, LLMs can solve more complex problems than when used alone \citep{cobbe2021training, lightman2024lets}. This is particularly true in real deployment settings, when models encounter new, verifiable reasoning questions which they must solve without access to ground-truth answers.

\looseness=-10000
Despite the ubiquity of solution verification, studies of solver–verifier interactions have remained limited in scope. Prior work has largely examined how a single model verifies its own solutions (self-verification) and improves its own solutions (self-improvement) \citep{song2025mind}. However, self-verification is not guaranteed to be effective. Models may be biased toward their own reasoning patterns, and their training may reinforce these tendencies. Additionally, this focus on self-verification offers little insight into verification performance when the solver and verifier differ. We therefore broaden our analysis to additionally include both intra-family and cross-family verification and ask the following central question:
\begin{quote}
    \vspace{-0.05in}
    \looseness=-10000
    \it
    When does verification actually pay off, and how do factors such as model family, model size, reasoning post-training, solver–verifier similarity, and task type influence its effectiveness?
\end{quote}

\looseness=-10000
To accomplish this, we evaluate verifiers across a diverse suite of tasks, including synthetic tasks used to test precise logical reasoning or symbolic computation (3SAT, Sudoku, and Matrix Multiplication), mathematical reasoning tasks (AIME \citep{AIME2025}, GSM8K \citep{cobbe2021training}), commonsense and factual reasoning (CSQA \citep{csqa}, GPQA \citep{gpqa}), and broad domain knowledge (MMLU in STEM and in social sciences \citep{mmlu}) using 37 models from 7 model families. 
% The tasks we chose enable evaluation with precise, objectively correct labels, while also capturing important skills for solving reasoning problems in real deployment settings. 
The tasks we chose provide ground-truth labels needed to rigorously measure verifier quality, while also capturing a broad range of skills required in real deployment settings. 
Using open-source model families that have base and post-trained pairs, size variants, reproducible inference pipelines, and explicit reasoning traces, we can study verification systematically.  

\looseness=-10000
We find that self-verification does not always ``pay off.'' Models often favor solutions resembling their own reasoning (Sections~\ref{sec:bettersolverbetterverifiers} and \ref{sec:similaritycorrelation}) and reasoning post-training can sharpen this bias (Section~\ref{sec:posttraining}), harming performance during self-verification or intra-family verification. Yet, this same bias makes stronger models more effective as cross-family verifiers, where the solver's distribution differs from their own. Additionally, we find that some tasks inherently benefit less from verification than others (Section~\ref{sec:datasetverifiability}). Therefore, we present the following contributions, which offer actionable and empirically supported guidance for how to use verifiers effectively.
\begin{itemize}[left=4pt,topsep=0pt]
    \item \looseness=-10000 \textbf{New Metric: Verifier Gain.}
    Verifier accuracy alone provides an incomplete picture of verifier usefulness at test time. To address this, we derive \textit{verifier gain}, a metric that simulates the improvement obtained from a verifier during test-time rejection sampling. We empirically study rejection sampling with verifiers and show that this theoretical formulation closely reflects empirical performance trends.
    \item \looseness=-10000 \textbf{Self-Improvement, Intra-Family Improvement, and Cross-Family Improvement.} We find that cross-family verification is often more beneficial than intra-family verification or self-verification, comparing particularly favorably to the latter. Looking deeper, we show that the verifier gain decreases as the solution distributions of solver and verifier become more similar. Furthermore, our results suggest that as models become stronger, whether through increased scale, post-training, or simply higher solver accuracy, they become less effective as self-verifiers and more effective as cross-family verifiers.
    \item \looseness=-10000 \textbf{Dataset Verifiability.} We study whether tasks that are easy to solve are also easy to verify and whether some tasks are inherently more verifiable than others. We find that verification accuracy generally correlates with solver accuracy, though self-verification yields little verifier gain across all tasks. We also observe that a clear subset of tasks involving mathematical or logical reasoning consistently produces higher verifier gains.
\end{itemize}

To the best of our knowledge, our work is the first systematic study of solver–verifier interactions across self-, intra-family, and cross-family regimes, spanning both base and reasoning post-trained models. Additionally, we introduce and validate verifier gain as a lightweight predictor of rejection-sampling improvements, examine the effects of reasoning post-training on verification, and connect observed differences in verification performance to output-distribution similarity between solver and verifier.

\clearpage

\section{Related Work}

\looseness=-10000
\textbf{Verifiers.} \citet{weng2023-better-reasoners}, \citet{wu-etal-2024-key-condition-verification}, and \citet{jiang-etal-2024-forward-backward} develop outcome-level self-verification methods by predicting parts of the question conditioned on the solution. To reduce hallucinations, \citet{dhuliawala-etal-2024-chain-of-verification} have language models fact-check their own generations by generating fact-check questions. Researchers have also trained general-purpose outcome-level verifiers~\citep{hosseini2024vstar, zhang-generativeverifiers, cobbe2021training} and value models~\citep{yu-etal-2024-ovm}, either independently or jointly with the solver~\citep{shen-etal-2021-generate-rank, sareen2025putting-value}. Unlike these trained outcome reward models, our work studies off-the-shelf LLMs prompted to verify solutions, which requires no task-specific training data or finetuning. \citet{song2025mind} investigate the performance improvement caused by using an outcome-level verifier (the GV-Gap), and how this improvement changes as the solver or verifier increases in capacity. However, they primarily focus on cases where the solver and verifier are the same model. Additionally, they only study base models and do not consider post-trained models in their analysis. We finally note concurrent work by \citet{zhou2025variation}, which also studies the factors that influence test-time verification. In contrast to our work, they focus on the effects of problem difficulty and generator capability, and do not investigate the effects of model family or reasoning post-training. Finally, verifiers also have their limitations, potentially producing false positives~\citep{stroebl2024inference-flaws}, eliminating valid reasoning paths~\citep{yu2025scaling-flaws-math}, and failing to select the right solution~\citep{brown2024large}. 

\looseness=-10000
\textbf{Scaling test-time compute.} A simple method for scaling test-time compute involves sampling several candidates and selecting one \textit{post hoc}, for example via Best-of-$N$. This can take the form of sample-and-rank approaches \citep{nichols2020collaborative}, majority vote \citep{wang2023selfconsistency}, model-based aggregation \citep{chen2024universal}, or sampling then filtering \citep{weng2023-better-reasoners}. Recent work has focused on studying scaling test-time compute with verifiers. For example, \citet{zhao2025sample} study random sampling with self-verification, \citet{chen2025sets} study combining parallel sampling with self-correction, and \citet{singhi2025solve} compare Self-Consistency~\citep{wang2023selfconsistency} to scaling with a generative verifier. \citet{snell2025scaling-tt-compute} investigate compute-optimal approaches to test-time scaling with process-level verification. 

\looseness=-10000
\textbf{Self-improvement.} Researchers have also studied LLM self-improvement and self-evaluation, with some voicing skepticism \citep{huang2024large-cannot-self-correct, kamoi-etal-2024-when-self-correct, olausson2023self, llm-evaluators, stechly2025on}. On the other hand, recent work has provided a theoretical framework for self-improvement via distribution sharpening, along with empirical support \citep{huang2024self-sharpening}. \citet{zhang-etal-2024-small-large-ver} look specifically at self-improvement for small models, arguing that they need to be paired with a stronger verifier. Some practical methods for self-improvement use natural language feedback \citep{madaan2023selfrefine, shinn2023reflexion, kim2023language} or train models for self-correction explicitly \citep{welleck2023generating}. Other methods use tools \citep{gou2024critic}, particularly code interpreters \citep{zhou2024solving}, to iteratively improve solutions. 

\begin{figure*}[t]
    \centering
    \includegraphics[width=\textwidth]{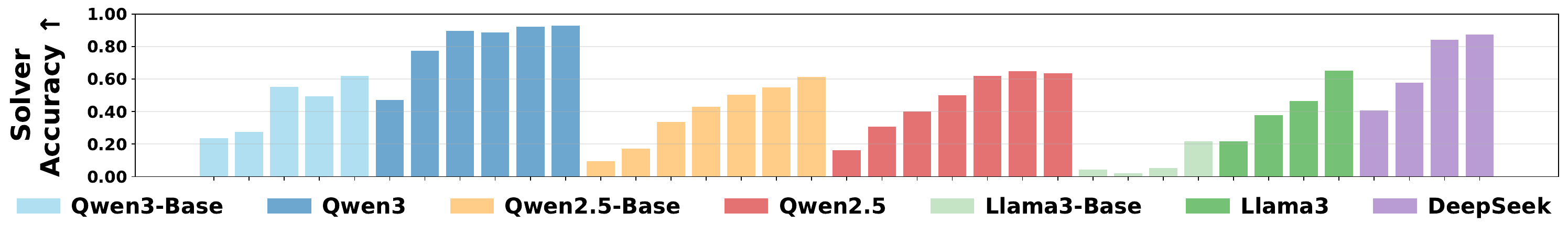}
    \caption{
        \looseness=-10000
        Average solver accuracy of each model over all datasets. Base model families are suffixed by \textbf{-Base}. Models within each family are ordered in increasing size. We show information for each evaluated model in Table~\ref{tab:allmodels}. 
    }
    \label{fig:cross-dataset-solver}
\end{figure*}

\section{Preliminaries}\label{sec:preliminaries}

\looseness=-10000
In this section, we establish the framework used throughout this work. We define datasets, solvers, and verifiers, introduce the metrics used to evaluate solver and verifier behaviors, and specify the verification settings in our empirical analysis.

\subsection{Datasets, Solvers, and Verifiers}

\looseness=-10000
Let $\mathcal{D}\subseteq \mathcal{X}\times\mathcal{Y}^\star$ be a dataset of pairs $(x,\mathcal{Y}_x)$, where $x\in\mathcal{X}$ is a problem and $\mathcal{Y}_x\subseteq \mathcal{Y}$ is a non-empty set of correct solutions.
A solver $S:\mathcal{X}\to \mathcal{Y}$ is an LLM that produces a solution $y$ for a given problem $x$, and a verifier $V:\mathcal{X}\times\mathcal{Y}\to\{0,1\}$ is an LLM that evaluates a problem–solution pair and returns a binary judgment. Following \citet{song2025mind}, who find chain-of-thought (CoT) verification more stable than multiple-choice formats, we instruct both solvers and verifiers to generate CoT reasoning before producing their final answers and judgments. We define the correctness indicator as $c(x,y) = \mathbbm{1}\{y \in \mathcal{Y}_x\}$.

\subsection{Evaluation Metrics}\label{sec:evalmetrics}

\looseness=-10000
We define the accuracy of a solver $S$ on a dataset $\mathcal{D}$ as the expected correctness of its outputs over all problems in the dataset: $\mathbb{E}_{(x,\mathcal{Y}_x)\sim \mathcal{D},\, y\sim S(x)}\big[\,c(x,y)\,\big]$. Verifier performance has several dimensions. We report the verifier accuracy, F1-Score, and precision (whose definitions are included in Appendix~\ref{detailverifiermetric} for reference) for our verification settings. In addition, we compute the true positive rate (TPR), false positive rate (FPR), and false negative rate (FNR):

\begin{center}
\scalebox{1.0}{$ %scaled to match equation below
    \begin{aligned}
        \text{TPR}(S,V;\mathcal{D}) &= \mathbb{E}[\,V(x,y) \mid y\in \mathcal{Y}_x\,] \\
        \text{FPR}(S,V;\mathcal{D}) &= \mathbb{E}[\,V(x,y) \mid y\notin \mathcal{Y}_x\,] \\
        \text{FNR}(S,V;\mathcal{D}) &= \mathbb{E}[\,1-V(x,y) \mid y\in \mathcal{Y}_x\,] = 1 - \text{TPR}(S,V;\mathcal{D}).
    \end{aligned}
$}
\end{center}

\looseness=-10000
Our primary goal is to evaluate whether using a verifier $V$ can improve a solver $S$ at test time via rejection sampling, where solver outputs are repeatedly sampled until the verifier accepts one. Assuming the solver has a non-zero probability of sampling a correct solution, in the limit of infinite resampling, the expected correctness of the accepted solution converges to the verifier’s precision, or the proportion of accepted solutions that are actually correct. To quantify the improvement from combining a solver with a verifier, we define \textbf{verifier gain}:

\begin{equation}\label{eqn:verifiergain}
\scalebox{1.0}{$ %scaled so that number fits on the same line
    \text{Gain}(S,V;\mathcal{D}) = \text{Precision}(S,V;\mathcal{D}) - \text{SolverAcc}(S;\mathcal{D}).
$}
\end{equation}

\looseness=-10000
This is a simple, yet useful, metric because it quantifies the improvement in correctness induced by filtering with the verifier. TPR/FPR describe the verifier in isolation; solver accuracy describes the generator in isolation, but verifier gain measures the benefit of the combined system. Since this is an asymptotic metric, it serves as a bound on the improvement attainable by verifier-based rejection sampling. Throughout this work, we use verifier gain to compare differences in verification behavior.

\subsection{Verification Settings}

\looseness=-10000
We group models into families (e.g., \texttt{Llama3}, \texttt{Qwen2.5}), where each family contains related models of varying sizes. Because base and post-trained models often exhibit substantially different behaviors, we treat them as distinct families. For example, the base model \texttt{meta-llama/Llama-3.1-70B} belongs to the \texttt{Llama3-Base} family and the post-trained model \texttt{meta-llama/Llama-3.1-8B-Instruct} belongs to the \texttt{Llama3} family. We study three different verification settings, defined by the solver-verifier relationship:

\looseness=-10000
\begin{itemize}[left=4pt]
    \item \looseness=-10000 \textbf{Self-Verification.} The solver and verifier are the same model, so the model verifies its own solutions. For example, when a 70B \texttt{Llama3} model is used as both the solver and the verifier, the verification metric (e.g., accuracy, FPR) is computed on this single pairing.
    \item \looseness=-10000 \textbf{Intra-Family Verification.} The verifier evaluates solutions produced by other models within the same family. For example, a 70B \texttt{Llama3} verifier may evaluate outputs from 8B or 13B \texttt{Llama3} solvers. The reported metric is averaged over all such within-family solvers, excluding the self-verification case.
    \item \looseness=-10000 \textbf{Cross-Family Verification.} The verifier evaluates solutions from models of different families. For example, a base \texttt{Llama3} verifier may evaluate outputs from \texttt{Qwen3} or from a post-trained \texttt{Llama3}. The reported metric is averaged over all such cross-family solvers.
\end{itemize}

\looseness=-10000
During our evaluation, we partition our computed verifier metrics according to the verification setting and average within each partition.

\section{Experimental Setup}\label{sec:experimentalsetup}

\looseness=-10000
\paragraph{Models.} We evaluate the solver and verifier abilities of 21 post-trained models from the \texttt{Llama3}~\citep{llama3}, \texttt{Qwen2.5}~\citep{qwen2.5}, \texttt{Qwen3}~\citep{qwen3}, and \texttt{DeepSeek-R1}~\citep{deepseek} families. For our study of reasoning post-training effects in Section~\ref{sec:posttraining}, we additionally evaluate 12 base models from the \texttt{Qwen2.5-Base} and \texttt{Qwen3-Base} families. 
% We specifically choose off-the-shelf models which can act as both solvers and verifiers for our experiments. 
We specifically choose off-the-shelf, general-purpose models so that each model can serve as both a solver and a verifier, allowing controlled comparison across these roles.
Model sizes range from 0.5B to 72B parameters. The full model list, with sizes, families, and HuggingFace identifiers, is provided in Appendix~\ref{modeldetails}. The legend of Figure~\ref{fig:cross-dataset-solver} displays the seven model families and the color scheme assigned to each.

\looseness=-10000
\paragraph{Datasets.} We compile a broad suite of real-world and synthetic tasks spanning diverse domains, including tasks requiring mathematical reasoning (GSM8K, AIME), commonsense knowledge (CSQA), and domain-specific factual knowledge of varying breadth (MMLU (STEM), MMLU (Social Sciences), and GPQA). We also construct synthetic tasks to assess logical reasoning (3SAT), structured puzzle solving (Sudoku), and symbolic computation (Matrix Multiplication). Further details with synthetic task examples are provided in Appendix~\ref{datasetdetails}. 

\looseness=-10000
\paragraph{Evaluation.} Datasets like Matrix Multiplication and the natural-language benchmarks contain a single ground-truth answer per problem, so we extract boxed solver outputs and evaluate via exact matching. In contrast, datasets like Sudoku and 3SAT may admit multiple valid solutions, so we evaluate solver outputs according to each task's rules. To evaluate verifiers, we give each model the problem and solver answer to generate CoT reasoning before producing a boxed ``correct" or ``incorrect" as the final judgment.

\looseness=-10000
\paragraph{Implementation.} For both solvers and verifiers, we generate with temperature~$0.7$, top-p~$0.9$, and a maximum output length of 8192 tokens. We discard outputs that do not contain a boxed answer. All inference experiments are run using vLLM on H200 GPUs. Prompts and additional details on output filtering are provided in Appendix~\ref{experimentalsetupdetails}.

\looseness=-10000
Upon publication, we plan to open-source all experiment and data generation code.

\section{Results}

\subsection{Does Verifier Gain Predict Improvements from Resampling?}\label{sec:verifiergain}

\begin{wrapfigure}[14]{r}{0.5\textwidth}
    \centering
    \includegraphics[width=\linewidth]{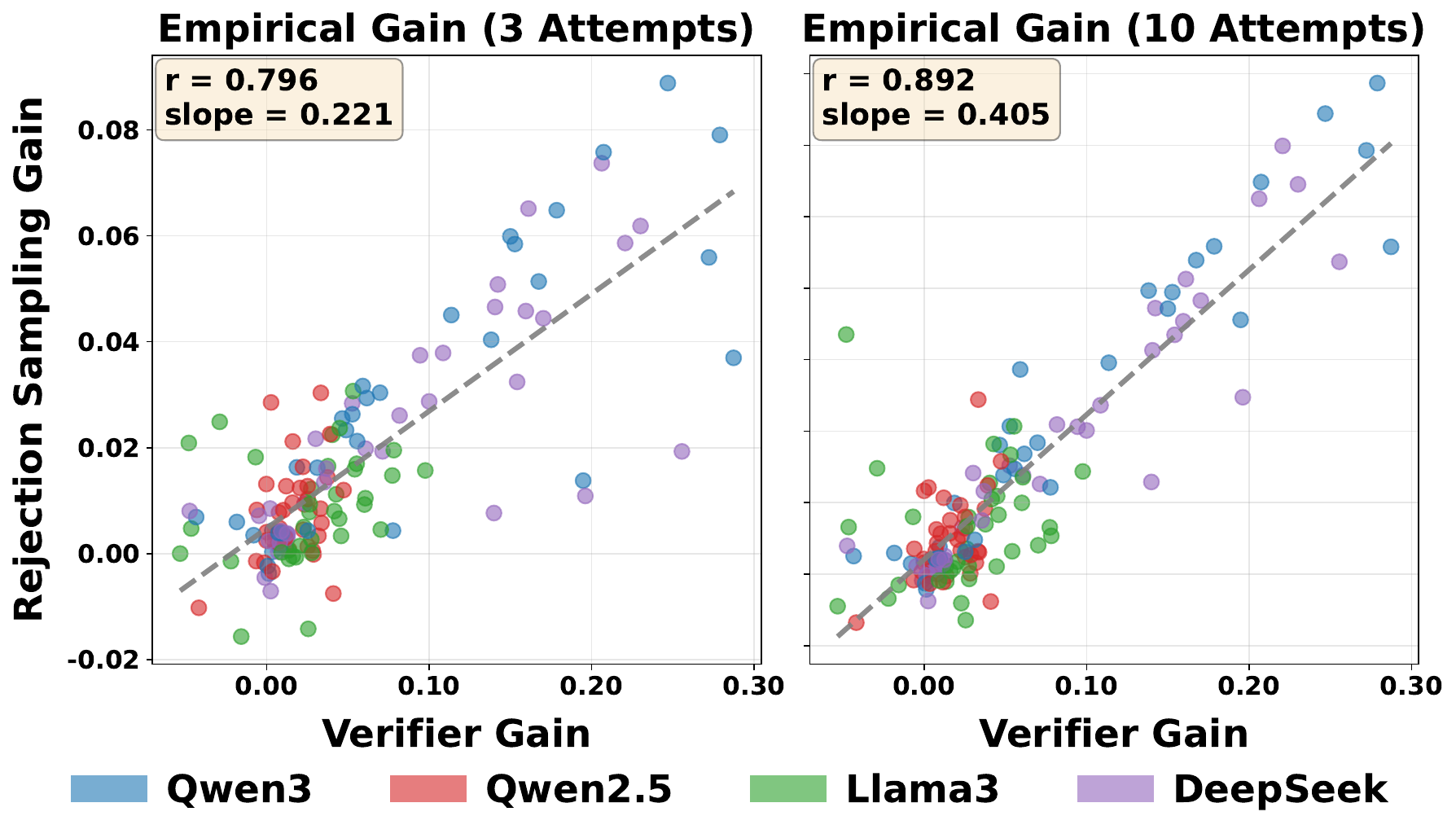}
    \vspace{-0.25in}
    \caption{
        Verifier gain (Equation~\ref{eqn:verifiergain}) predicts empirical rejection sampling gain. Each point is one solver--verifier pair averaged across datasets, colored by verifier family.
    }
    \label{fig:cross-dataset-verifier-gap}
\end{wrapfigure}

\looseness=-10000
We first empirically validate our verifier gain metric, which estimates the expected improvement in a solver's accuracy when using a verifier for rejection sampling. To assess how well this metric predicts real performance, we conduct rejection sampling experiments across all $12 \times 12$ solver–verifier pairs from a 12-model subset of our post-trained models, consisting of the three smallest models from each of the four post-training families. For each problem, the solver generates solutions until the verifier labels one as correct, up to ten attempts. If no such solution is found, we retain the final attempt. 

Figure~\ref{fig:cross-dataset-verifier-gap} plots the empirical gain against the theoretical verifier gain for each pair, averaged across datasets. The Pearson correlation ($r$) strengthens considerably from 3 to 10 attempts, consistent with verifier gain being an asymptotic metric that better predicts performance as the resampling attempts grow. 

\begin{quote}
    \looseness=-10000
    \textbf{Takeaway:}
    Verifier gain reliably predicts rejection sampling gains and serves as a powerful comparative metric for evaluating solver–verifier pairs. Crucially, it can be estimated from a single verification round without costly rejection sampling experiments.
\end{quote}

\subsection{Do Better Solvers Make Better Verifiers?}\label{sec:bettersolverbetterverifiers}

\begin{wrapfigure}[24]{r}{0.5\textwidth}
    \vspace{-0.35in}
    \centering
    \includegraphics[width=0.975\linewidth]{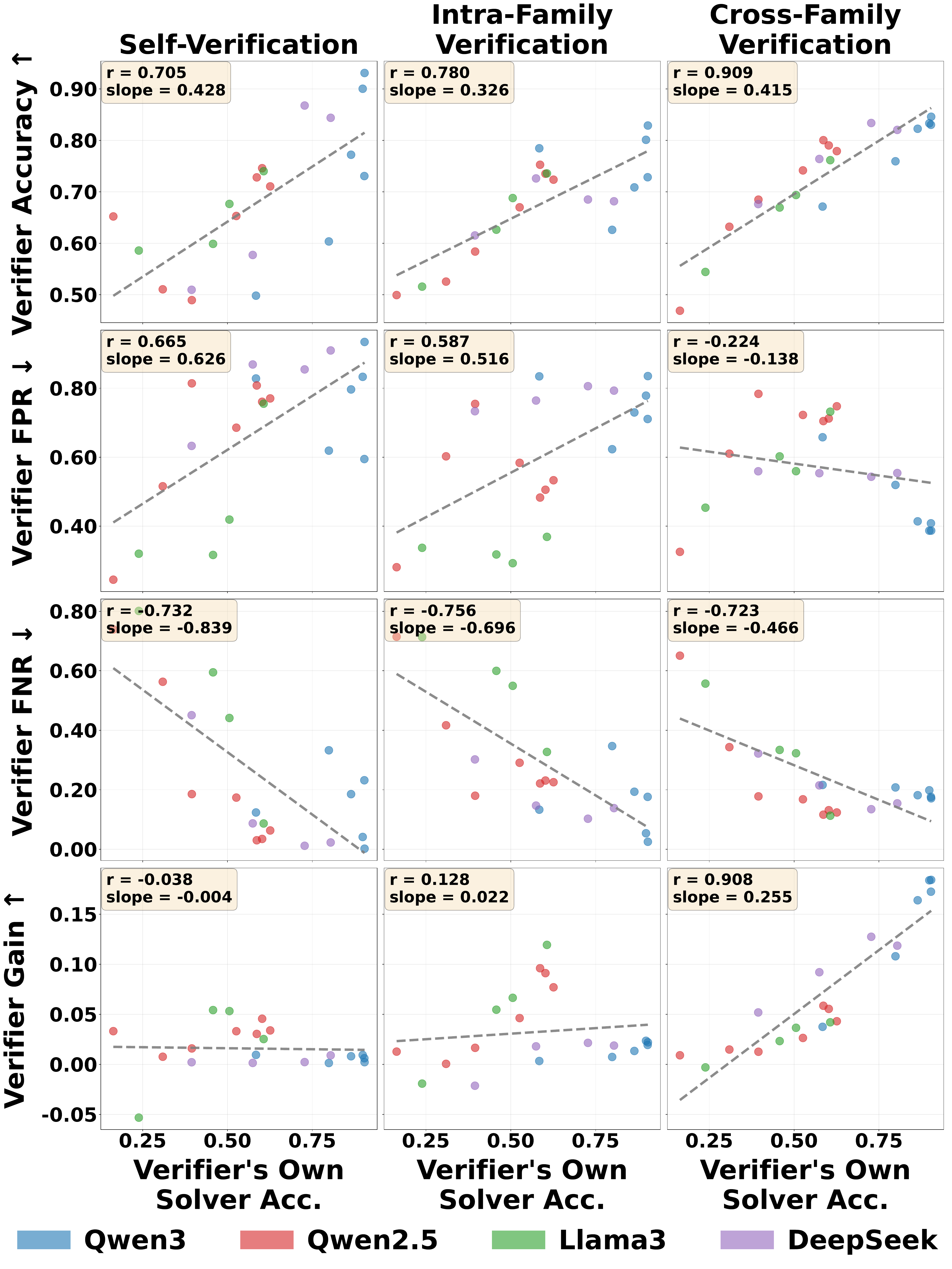}
    \vspace{-0.15in}
    \caption{
        \looseness=-10000
        Correlation between each verifier's metrics (rows) and its own solver accuracy for all 21 post-trained models, averaged over all datasets. Each metric is computed over our three verification settings (columns).
    }
    \label{fig:cross-dataset-verifier-scatter-solver-acc}
\end{wrapfigure}

We next study the relationship between solver and verifier performance, and how to best measure the benefits of verification.

\paragraph{Solver performance.} 
\looseness=-10000
We benchmark all 37 models on each of our 9 datasets, averaging performance across tasks (Figure~\ref{fig:cross-dataset-solver}) and reporting task-level results in Appendix~\ref{appendix:solver-acc}. Overall, solver accuracy increases with model capacity. Models in the \texttt{Qwen3} and \texttt{DeepSeek} families perform particularly well, whereas \texttt{Llama3-Base} performs poorly due to base models being unfamiliar with the question–answering instruction format. Within each family, we observe clear performance scaling for \texttt{Qwen2.5-Base} and \texttt{DeepSeek}, with the remaining families showing similar upward trends.

\paragraph{Correlating verifier and solver performance.} 

\looseness=-10000
After establishing solver accuracy, we analyze whether a model’s solver performance correlates with its performance as a verifier (Figure~\ref{fig:cross-dataset-verifier-scatter-solver-acc}). For each of our 21 post-trained models and each dataset, we evaluate verification on the same set of solver models to obtain verifier accuracy, F1-Score, precision, FPR, FNR, and gain for every solver–verifier pair. For each verifier, we then partition the results by our three verification settings and average within each setting over solvers and datasets.

\looseness=-10000
While verifier accuracy tends to improve with the verifier’s own solver accuracy, the relationship becomes more nuanced when examining other metrics. The FPR increases during self-verification and intra-family verification but decreases slightly during cross-family verification, indicating that when a strong solver is used as a verifier, it is more likely to falsely label a solution as correct if it was generated by itself or another model in its family. We provide additional visualizations for F1-Score and precision in Appendix~\ref{appendix:f1_and_precision}.

\begin{figure*}[t]
    \centering
    \includegraphics[width=\textwidth]{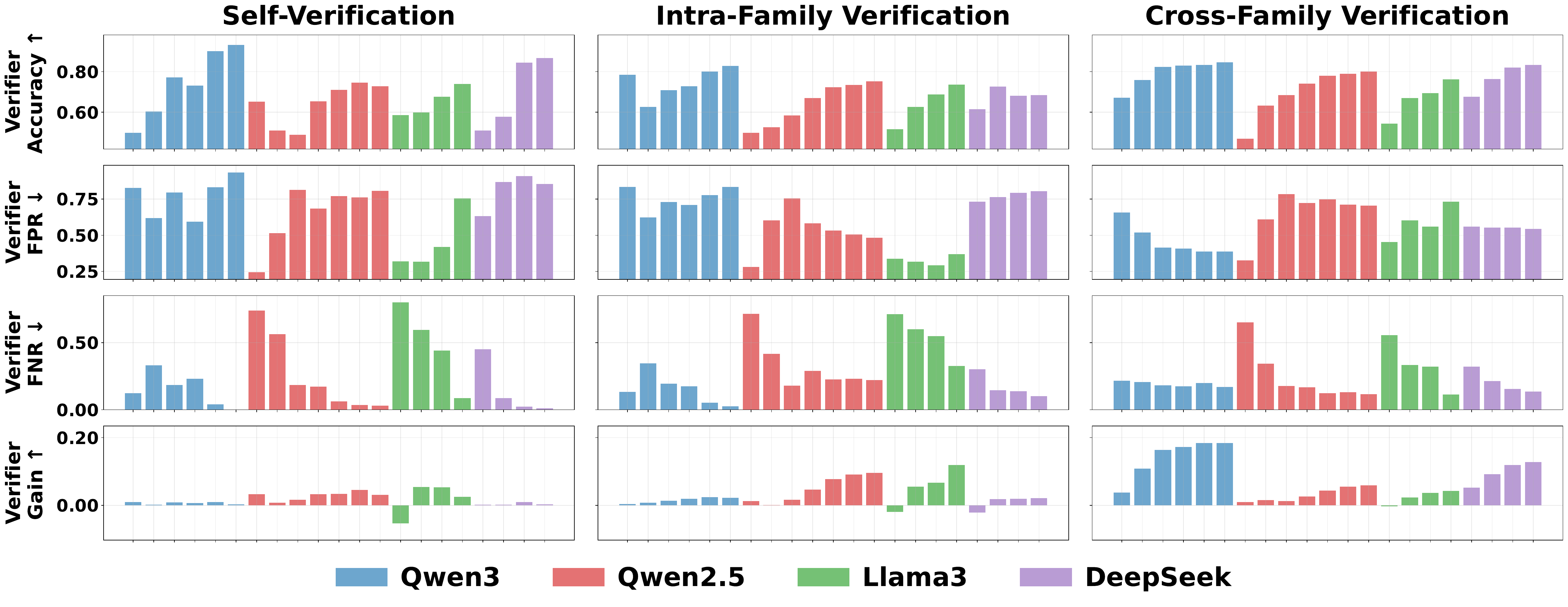}
    \caption{
        \looseness=-10000
        We show each verifier's metric (rows) against model size for all 21 post-trained models, averaged over all datasets. Models are separated by family and ordered by increasing size. Each metric is computed over our three verification settings (columns).
    }
    \label{fig:cross-dataset-verifier-barplots}
\end{figure*}

\looseness=-10000
To better interpret the trends suggested by the accuracy and FPR results, we examine verifier gain in the final row. This visualization offers a clearer view of verification quality: self-verification yields the smallest gains, and more accurate solvers do not exhibit greater self-improvement. Gains increase slightly in intra-family verification, while cross-family verification provides the greatest potential benefits.
% Verifier gain quantifies the expected benefit of using the verifier during rejection sampling, i.e., repeatedly sampling until the verifier accepts a solution (a common solver–verifier interaction setting \citep{song2025mind}).

\paragraph{Examining verifier performance at different model families and sizes.} 

\looseness=-10000
In Figure~\ref{fig:cross-dataset-verifier-barplots}, we repeat the experiments from Figure~\ref{fig:cross-dataset-verifier-scatter-solver-acc} but plot each verifier metric as a function of model size within each model family. Verification accuracy and FNR consistently improve as models become larger, whereas FPR behaves more inconsistently, often increasing with model size (e.g., intra-family verification for \texttt{DeepSeek}).

\looseness=-10000
For state-of-the-art reasoning post-trained models such as \texttt{Qwen3} and \texttt{DeepSeek}, we find that verifier gains are largest in the cross-family setting, smaller in the intra-family setting, and minimal during self-verification. At first glance, this appears to contradict \citet{song2025mind}, who report that self-verification GV-Gaps increase with more pre-training FLOPs. However, their analysis focuses on older model families, and Figure~\ref{fig:cross-dataset-verifier-scatter-solver-acc} likewise shows larger verifier gains for older models such as \texttt{Qwen2.5} and \texttt{Llama3}.

\looseness=-10000
We hypothesize that stronger reasoning post-trained models like \texttt{DeepSeek} and \texttt{Qwen3} show negligible gains in self-verification and limited gains in intra-family verification because they may already engage in \textit{spontaneous} self-verification when used as solvers, reducing the benefit of an additional \textit{forced} verification round. To test this, we measure the spontaneous self-verification rate of all 21 post-trained models across all 9 datasets by scanning solver outputs for self-verification keywords (e.g., ``let me check,'' ``wait,'' ``reconsider,'' ``that's wrong''; full list in Appendix~\ref{appendix:keywords}). Indeed, \texttt{Qwen3} and \texttt{DeepSeek} self-verify in 96\% and 73\% of outputs, respectively, while \texttt{Llama3} and \texttt{Qwen2.5} self-verify in only 1--2\%. This aligns with our hypothesis: models that already spontaneously self-verify during solving leave little room for an additional verification pass to add value. More broadly, it suggests the relevant question is whether reasoning post-training introduces spontaneous self-verification.

% We hypothesize that stronger post-trained models like \texttt{DeepSeek} and \texttt{Qwen3} show negligible gains in self-verification and limited gains in intra-family verification for two reasons: (a) they may already engage in \textit{spontaneous} self-verification when used as solvers, reducing the benefit of an additional \textit{forced} verification round, and (b) their distributions are significantly sharpened by post-training \citep{huang2024self-sharpening}, which limits the improvement obtained from rejection sampling.

\begin{quote}
    \textbf{Takeaways:}
    \begin{enumerate}
        \item \looseness=-10000 Verifier models are biased toward accepting incorrect solutions when performing self-verification or intra-family verification.
        \item \looseness=-10000 Verification accuracy alone is not a reliable predictor of how much a verifier can improve a solver at test time. Instead, computing verifier gain using solver accuracy and verifier precision provides a more reliable metric.
        \item \looseness=-10000 While model families like \texttt{Llama3} and \texttt{Qwen2.5} show some ability to self-improve, stronger model families like \texttt{DeepSeek} and \texttt{Qwen3} do not, which we find is linked to the latter already spontaneously self-verifying during solving (73--96\% vs.\ 1--2\%).
    \end{enumerate}
\end{quote}

\subsection{Are Verifiers Biased Toward Solutions That Resemble Their Own?}\label{sec:similaritycorrelation}

\begin{wrapfigure}[32]{r}{0.5\textwidth}
    \centering
    \includegraphics[width=1.0\linewidth]{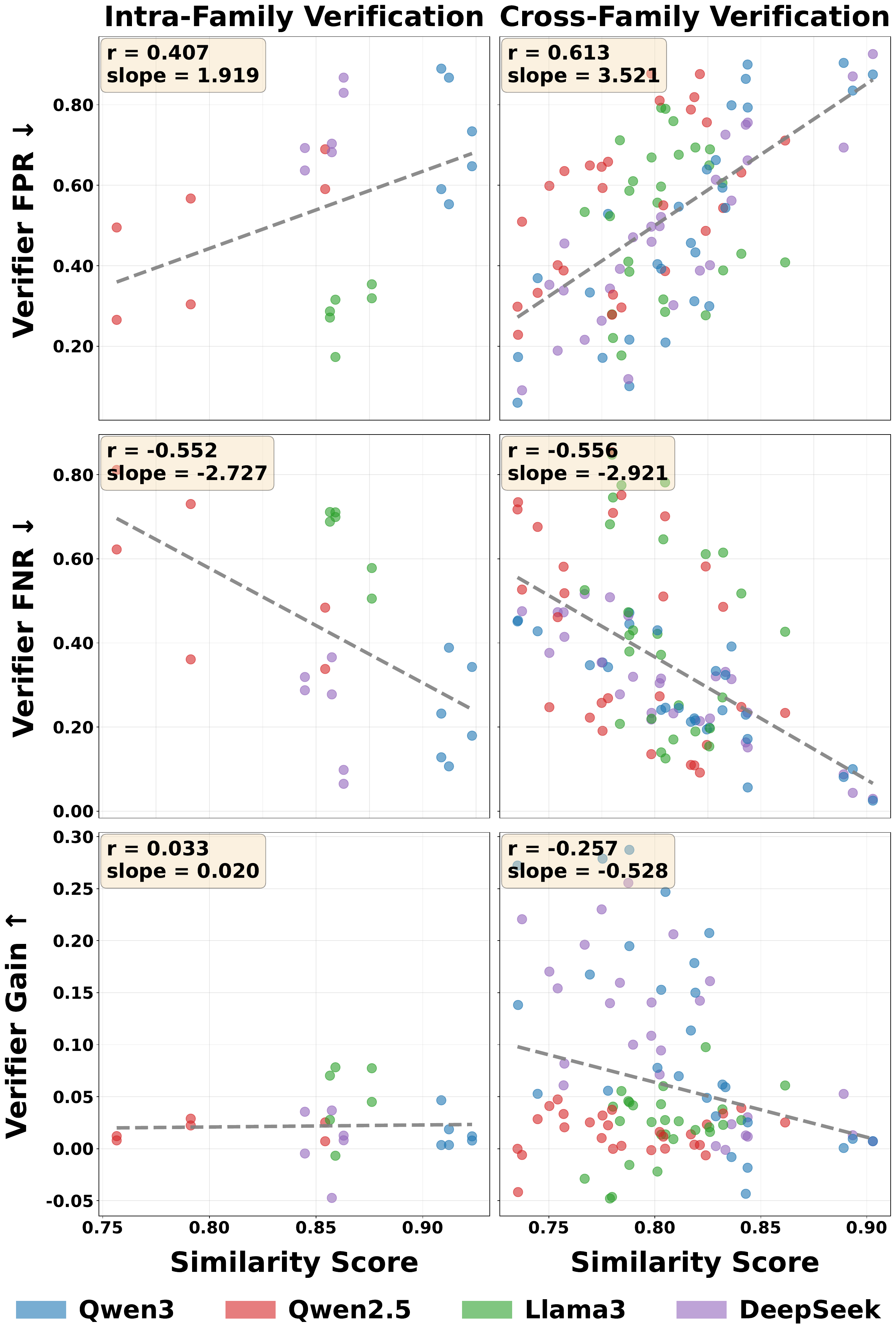}
    \vspace{-0.25in}
    \caption{
        \looseness=-10000
        Correlation between verifier metrics and similarity scores between solver-verifier pairs. Each marker is colored based on the verifier model family. Self-verification is omitted as it yields only one data point per model, too few for reliable correlation analysis.
    }
    \label{fig:similarity-plot}
\end{wrapfigure}

\looseness=-10000
Humans tend to judge solutions that resemble their own reasoning as more likely correct. This mirrors self-enhancement bias~\citep{self-enhancement-bias}, where individuals evaluate themselves more favorably than evidence suggests. Our results in Figures~\ref{fig:cross-dataset-verifier-scatter-solver-acc} and~\ref{fig:cross-dataset-verifier-barplots} show that strong reasoning models benefit the least from self-verification and the most from cross-family verification, suggesting an analogous effect in solver–verifier interactions.

\looseness=-10000
To directly investigate this, we study the relationship between verifier performance metrics and the \textbf{solver–verifier similarity score}, defined as the average cosine similarity between two models' solution embeddings across all dataset problems. We conduct cross-verification experiments using 12 post-trained models (the three smallest from each of four families) and compute all verifier metrics for each pair. For intra-family verification, each solver has 2 same-family verifiers (excluding itself), yielding $12 \times 2 = 24$ pairs. For cross-family verification, each solver has 9 verifiers from other families, giving $12 \times 9 = 108$ pairs. Solutions are embedded using \texttt{sentence-transformers} \texttt{/all-mpnet-base-v2}.

\looseness=-10000
Figure~\ref{fig:similarity-plot} shows that, for both intra-family and cross-family settings, the more similar the solver and verifier distributions, the more likely the verifier is to accept incorrect answers. While intra-family verifier gains are too small to yield a strong correlation, cross-family gains decrease significantly as similarity increases, indicating that choosing a dissimilar verifier leads to more reliable verification. We replicate this analysis using log-likelihood as an alternative similarity metric with the same directional findings, but we prefer cosine similarity as the primary metric because log-likelihood conflates distributional similarity with intrinsic predictability of the solver's text (see Appendix~\ref{appendix:likelihood} for details).

\begin{quote}
    \looseness=-10000
    \textbf{Takeaway:} Higher similarity between solver and verifier solution distributions increases the verifier’s tendency to accept incorrect solver outputs, reducing verifier gain. Using a verifier with a meaningfully different solution distribution mitigates this bias.
\end{quote}

\subsection{How Does Reasoning Post-Training Affect Solver and Verifier Performance?}\label{sec:posttraining}

\begin{wrapfigure}[20]{r}{0.5\textwidth}
    \centering
    \vspace{-0.25in}
    \includegraphics[width=1.0\linewidth]{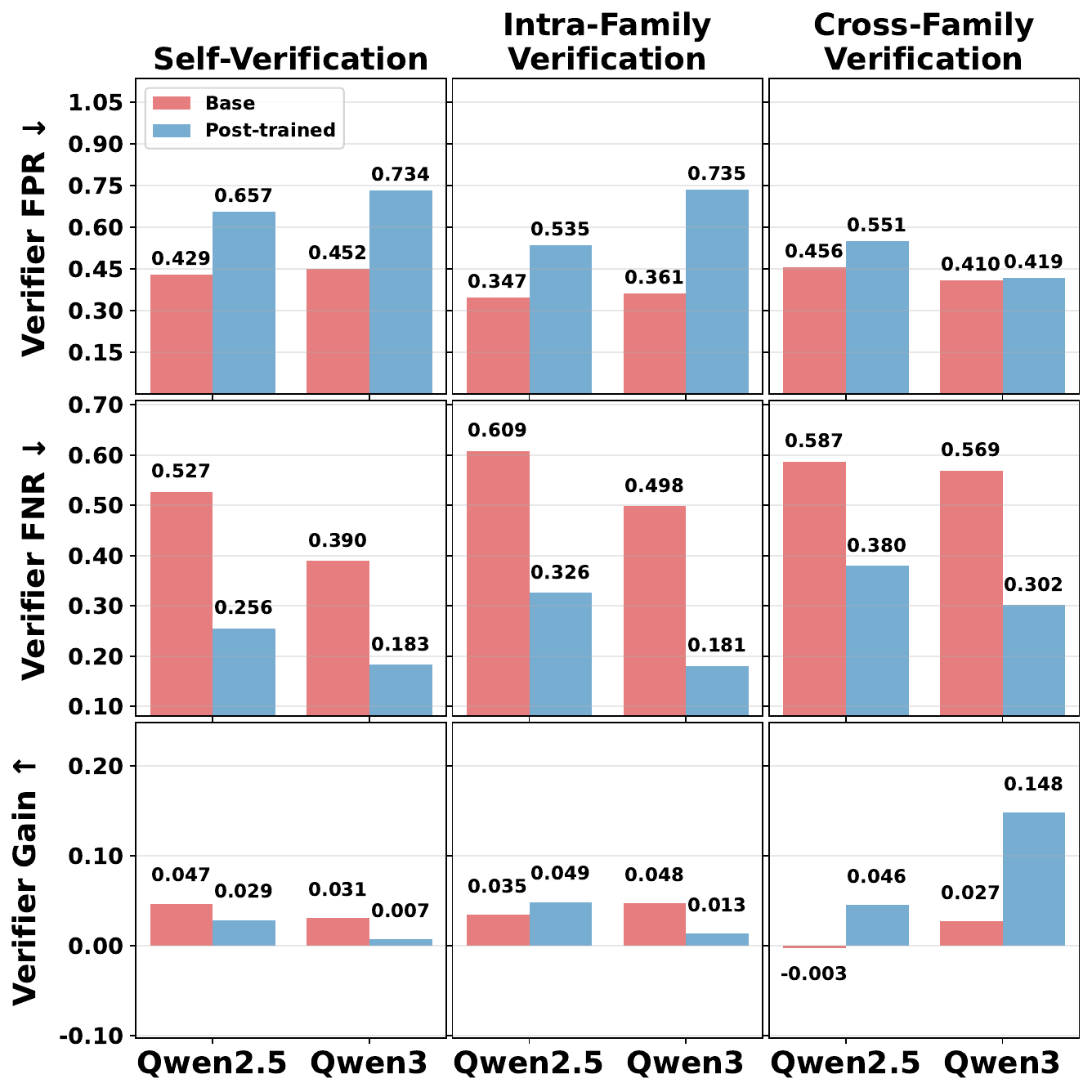}
    \vspace{-0.25in}
    \caption{
        Changes in verifier metrics of the \texttt{Qwen2.5-Base} and \texttt{Qwen3-Base} models from reasoning post-training.
    }
    \label{fig:post-train-verifier}
\end{wrapfigure}

\looseness=-10000
We examine how post-training influences verifier behavior, focusing on the \texttt{Qwen2.5-Base}/\texttt{Qwen2.5} and \texttt{Qwen3-Base}/\texttt{Qwen3} pairs, since other families are either too weak (\texttt{Llama3-Base}) or lack base models (\texttt{DeepSeek}). As both \texttt{Qwen2.5} and \texttt{Qwen3} use GRPO for reasoning post-training, our analysis primarily concerns its effects. Verification metrics are computed across all 37 base and post-trained models.

\looseness=-10000
\paragraph{Reasoning post-training's effect on solver performance.} We evaluate how solver accuracy changes after reasoning post-training, computing accuracy averaged across model families and datasets. As expected, post-training yields substantial improvements: \texttt{Qwen2.5} solvers improve by $8.2\%$ on average, while \texttt{Qwen3} solvers show a striking $35.4\%$ gain.

\paragraph{Reasoning post-training's effect on verifier performance.} 

\looseness=-10000
We next analyze how reasoning post-training affects verifier behavior (Figure~\ref{fig:post-train-verifier}). For each model, we compute verifier metrics against all solvers and datasets, partition by verification setting, and average within families. For both Qwen families, post-training increases FPR and reduces verifier gain in self-verification, despite improvements in FNR. Although \texttt{Qwen3} benefits more in solver accuracy (Figure~\ref{fig:post-train-solver}), its post-trained verifiers show higher FPRs and lower gains in both self- and intra-family verification. In contrast, both families, especially \texttt{Qwen3}, show substantial improvements in cross-family verification. Overall, reasoning post-training exacerbates trends identified previously, increasing false positives and limiting self-verification benefits.

\begin{quote}
    \looseness=-10000
    \textbf{Takeaway:} Reasoning post-training significantly improves problem-solving but can reduce self- and intra-family verification gains, while boosting cross-family verification performance.
\end{quote}

\subsection{How Does Task Type Affect Verifiability?}\label{sec:datasetverifiability}

\begin{figure*}
    \centering
    \includegraphics[width=\textwidth]{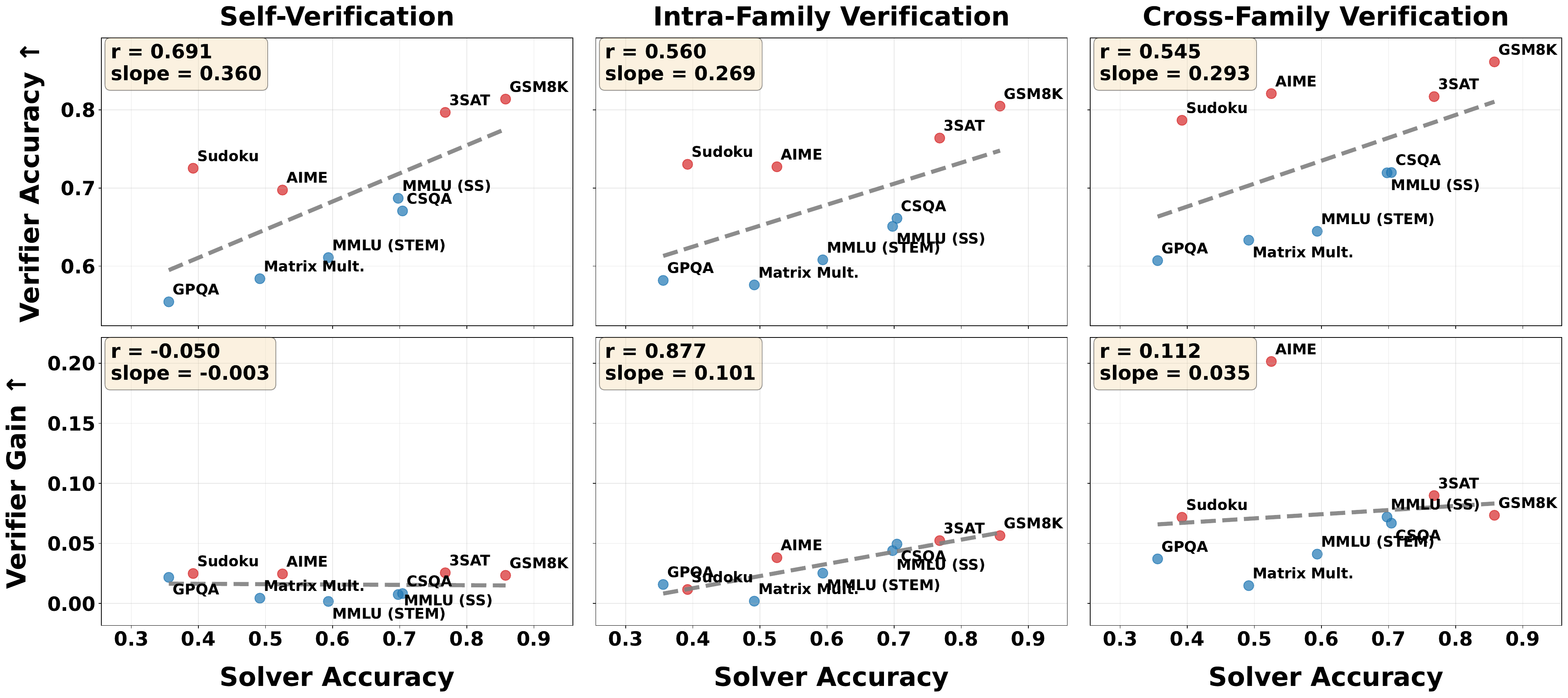}
    \caption{
    Correlation of verifier metrics (rows) with solver accuracies, averaged over solver-verifier pairs that belong to each verification setting (columns).}
    \label{fig:cross-dataset-task-scatterplots}
\end{figure*}

\looseness=-10000
Thus far, we have examined model-related factors. We now shift to a task-level perspective and ask: \textit{are tasks that are easy to solve also easy to verify?} In Figure~\ref{fig:cross-dataset-task-scatterplots}, we recompute the verifier metrics from Section~\ref{sec:bettersolverbetterverifiers}, average across all verifier models, and plot them against solver accuracies. Solver accuracy and verifier accuracy correlate strongly in all settings, but the picture is more mixed for verifier gains. During self-verification, we find essentially no correlation between verifier gain and solver accuracy, but a clear positive relationship emerges during intra-family and cross-family verification. Notably, AIME appears as an outlier, potentially because some models encountered similar problems during post-training.

\looseness=-10000
The best-fit lines for verifier accuracy reveal two distinct task clusters (colored red and blue), leading us to ask: \textit{are some tasks inherently easier to verify than others?} In Figure~\ref{fig:cross-dataset-task-scatterplots}, AIME, GSM8K, 3SAT, and Sudoku exhibit a higher ratio of verifier accuracy to solver accuracy and deliver higher gains across all settings. Sudoku and 3SAT require exponential solving time but allow polynomial-time verification, whereas verifying a matrix product offers no such shortcut. Among real-world datasets, GSM8K and AIME involve high-school-level mathematics, whereas MMLU (Social Sciences) requires domain-specific knowledge, CSQA relies on world knowledge, and GPQA and MMLU (STEM) draw on specialized natural science knowledge. For these latter tasks, verifying an answer requires essentially the same knowledge as solving it, confirming and extending \citet{song2025mind}'s finding that models cannot self-improve on factual recall tasks but can on Sudoku.

\begin{quote}
    \textbf{Takeaways:} 
    \begin{enumerate}
        \item \looseness=-10000 Tasks that are easy to solve tend to be easy to verify.
        \item \looseness=-10000 Tasks that are easier to solve tend to be more improvable through intra-family and cross-family verification, but not necessarily through self-verification.
        \item \looseness=-10000 Synthetic problems with logical or structured reasoning, as well as real-world tasks which rely on mathematical reasoning, are inherently easier to verify, and yield larger verifier gains, than those which require factual recall.
    \end{enumerate}
\end{quote}

\section{Conclusion}

\looseness=-10000
This work presents a comprehensive study of LLM-based verification for problem solving. We show that verification accuracy alone provides an incomplete picture, motivating \textit{verifier gain}, which measures the expected improvement from using a verifier for rejection sampling. We find lower verifier gains from self- and intra-family verification compared to cross-family, trends exacerbated by reasoning post-training and increasing model size. Further analysis reveals that these decreases correlate with greater similarity between solver and verifier solution distributions, indicating that verifiers are biased to accept solutions resembling their own. We also find that tasks involving logical or mathematical reasoning are inherently easier to verify than knowledge-recall tasks.    

Our results yield an actionable checklist for designing effective solver-verifier systems:
\looseness=-10000
\begin{itemize}[left=4pt,topsep=0pt,itemsep=0pt]
    \item \looseness=-10000 Use verifier gain, not accuracy, to evaluate a solver-verifier pair. Verification accuracy can be misleading (Section~\ref{sec:bettersolverbetterverifiers}), while verifier gain strongly predicts actual rejection sampling gains (Section~\ref{sec:verifiergain}).
    \item \looseness=-10000 Check whether the task is easier to verify than to solve. Logical and mathematical reasoning tasks yield higher verifier gains than knowledge-recall tasks (Section~\ref{sec:datasetverifiability}).
    \item \looseness=-10000 Prefer verifiers that "think differently" from the solver. Solution-distribution similarity increases false positives and reduces gains (Section~\ref{sec:similaritycorrelation}).
    \item \looseness=-10000 Avoid using strong reasoning models as their own verifiers due to their minimal self-verification gain from reasoning post-training (Sections~\ref{sec:bettersolverbetterverifiers},~\ref{sec:posttraining}).
\end{itemize}

\paragraph{Limitations and future work.} 

\looseness=-10000
Section~\ref{sec:similaritycorrelation} shows that LLMs are biased toward accepting incorrect solutions that resemble their own reasoning, indicating that it will be worthwhile to examine the origins of this bias in pre-training and/or post-training. Additionally, while we note that our solver-verifier setting corresponds to a number of real deployment settings, multi-turn problem solving is an increasingly popular way to solve difficult problems with LLM agents. A promising avenue for future research is to study verification in the multi-turn setting, where future interactions incorporate verifier feedback and also depend on previous solver outputs in the conversation.

\section*{Acknowledgement}
We thank members of the NYU Agentic Learning AI Lab for their helpful discussions. JL is supported by the NSERC PGS-D Scholarship. 
The work is supported in part by the Institute of Information \& Communications Technology Planning \& Evaluation (IITP) under grant RS-2024-00469482, funded by the Ministry of Science and ICT (MSIT) of the Republic of Korea in connection with the Global AI Frontier Lab International Collaborative Research. 
The compute is supported by the NYU High Performance Computing
resources, services, and staff expertise. We also thank Modal for providing additional compute resources.

\bibliographystyle{apalike}
\bibliography{ref}

\appendix

\section*{Appendix}

\section{Additional Details on Verifier Metrics}\label{detailverifiermetric}

\looseness=-10000
We show the mathematical definitions of relevant verifier metrics below. For clarity, we include dependencies (e.g., $(S,V;\mathcal{D})$) in the definitions, but sometimes omit them for brevity when the context is clear.

\begin{align*}
\text{VerifierAcc}(S,V;\mathcal{D}) &= \mathbb{E}_{(x,\mathcal{Y}_x)\sim \mathcal{D},\, y\sim S(x)}\big[\,\mathbbm{1}\{\,V(x,y) = c(x,y)\,\}\,\big] \\
\text{Precision}(S,V;\mathcal{D}) &= \mathbb{E}[\,c(x,y) \mid V(x,y)=1\,] 
= \frac{\text{SolverAcc}\cdot \text{TPR}}{\text{SolverAcc}\cdot \text{TPR} + 
(1-\text{SolverAcc}) \cdot \text{FPR}} \\
\text{Recall}(S,V;\mathcal{D}) &= \text{TPR} \\
\text{F1}(S,V;\mathcal{D}) &= \frac{2 \cdot \text{Precision} \cdot \text{Recall}}{\text{Precision} + \text{Recall}}
\end{align*}

\section{Additional Details on Models}\label{modeldetails}

\looseness=-10000
We show the information for each of our 37 evaluated models in Table~\ref{tab:allmodels}. 

\section{Additional Details on Datasets}\label{datasetdetails}

\subsection{Real-World Datasets}

\looseness=-10000
Note that for MMLU (STEM) and MMLU (Social Sciences), we concatenate questions from all subjects that belong to the STEM and Social Sciences supercategories in~\citet{mmlu}, respectively. 

\subsection{Synthetic Datasets}

\looseness=-10000
We generate three synthetic datasets, named 3SAT, Matrix Multiplication, and Sudoku, with 1000 samples each. We explain each synthetic dataset's generation parameters below.

\looseness=-10000
Each 3SAT CNF contains uniformly sampled numbers of variables and clauses from 2 to 8 (inclusive). Each Sudoku puzzle is a 9x9 grid with 12 randomly missing cells. Each Matrix Multiplication problem is about multiplying 2 4x4 integer matrices with values uniformly sampled from $[-5, 5]$. All data are generated in a way that ensures the existence of a valid solution. Note that while Matrix Multiplication has a singular correct answer for each problem, Sudoku and 3SAT are allowed multiple correct answers as long as the solver's answer is correct by their rules. 

\looseness=-10000
The generation code files for all synthetic datasets are seeded for reproducibility. 

\looseness=-10000
An example of a generated 3SAT problem:
\begin{verbatim}
## Problem Definition

**SAT (Boolean Satisfiability Problem)** is a fundamental problem in computer science 
where we need to determine if there exists an assignment of Boolean values (True/False) 
to variables that makes a given Boolean formula evaluate to True.
**Variables**: In this problem, variables are named as single letters. Each variable can 
be assigned either True (T) or False (F).
**Literals**: A literal is either a variable (like a) or its negation (like ~a, meaning 
"not a"). If a is True, then ~a is False, and vice versa.
**Clauses**: A clause is a disjunction (OR operation) of literals. A clause is satisfied 
(True) if at least one of its literals is True. For example, the clause (a or ~b) is True if 
either a is True OR b is False (or both).
**CNF (Conjunctive Normal Form)**: The Boolean formula is given in CNF, which is a 
conjunction (AND operation) of multiple clauses. The entire formula is satisfied only if 
ALL clauses are satisfied simultaneously.
**3SAT**: This is a special case of SAT where every clause contains exactly 3 literals.

## The Problem

Find a satisfying assignment for the following CNF formula: (~c or ~b or d) and 
(d or ~b or ~c) and (d or a or c) and (~c or d or a) and (b or ~a or d) and (c or d or ~b)

## Instructions

Provide your answer as a list of variable assignments, one per line, in the format 
"variable_name T" or "variable_name F." For example:
\boxed{
a T
b F
}
This means a=True, b=False.

Another example answer is
\boxed{
a F
b T
}
This means a=False, b=True.

Output and only output the T/F values for the variables that appear in the provided 
CNF formula.
\end{verbatim}

An example of a generated Sudoku problem:
\begin{verbatim}
## Sudoku Problem

**Sudoku** is a logic-based number-placement puzzle. The objective is to fill a 9x9 grid 
with numbers so that each column, each row, and each of the 3x3 sub-grids contains all 
of the numbers from 1 to 9.

## The Puzzle

Complete the following 9x9 Sudoku grid (empty cells are marked with '_'):

7 4 2 1 _ 5 8 9 6
1 6 9 2 4 8 3 5 7
8 5 3 _ _ 7 2 1 4
2 _ 8 9 7 1 4 6 5
5 7 6 4 8 2 9 3 _
4 9 1 3 _ 6 _ 8 _
3 1 5 8 2 4 6 7 9
6 8 _ 7 1 _ 5 2 3
_ 2 7 5 6 _ 1 4 8

## Instructions

Provide your answer as a completed 9x9 grid with all numbers filled in, formatted exactly 
like the puzzle above but with numbers instead of underscores.

For example, a completed 4x4 grid should look like:
\boxed{
1 2 3 4
3 4 1 2
2 3 4 1
4 1 2 3
}
\end{verbatim}

\looseness=-10000
An example of a generated Matrix Multiplication problem:
\begin{verbatim}
## Matrix Multiplication Problem

**Matrix Multiplication** is a fundamental operation in linear algebra where we compute 
the product of two matrices. For two square matrices A and B of size 4x4, the product 
C = A x B is computed as:

C[i][j] = Sum(k=0 to 3) A[i][k] x B[k][j]

## The Problem

Compute the product of the following two 4x4 matrices:

**Matrix A:**
0 1 1 4
-1 3 4 4
-2 -5 -5 0
-4 4 5 0

**Matrix B:**
1 2 0 5
1 -2 0 0
3 -1 -3 -3
2 5 -4 2

## Instructions

Provide your answer as the resulting 4x4 matrix C = A x B, formatted with each row
on a separate line and numbers separated by spaces.

For example, a 2x2 result matrix is formatted like:
\boxed{
1 2
3 4
}
\end{verbatim}

\section{Additional Details on Experimental Setup}\label{experimentalsetupdetails}

\looseness=-10000
We use the following solver prompt for all models:
\begin{verbatim}
Please reason step by step, and put your final answer within \boxed{{}}.

{question}
\end{verbatim}

We use the following verifier prompt for all models:
\begin{verbatim}
You are a teacher that is evaluating a student's answer to a question. 
Your task is to determine whether the answer is correct or incorrect.

Question: {question}

Student's Answer: {response}

Please evaluate the student's answer carefully. Consider:
- Is the answer factually accurate?
- Is the reasoning sound and logical?
- Does it fully address the question asked?

After your evaluation, provide your judgment in the 
following format:
- If the answer is correct, write: \boxed{{correct}}.
- If the answer is incorrect, write: \boxed{{incorrect}}.

First explain your analysis over the student's answer, then provide your final judgment in 
the boxed format. Make sure the final judgment is either "correct" or "incorrect" inside 
the \boxed{{}}. Do not put anything else in \boxed{{}}. Do not repeat the student's answer 
in \boxed{{}}.
\end{verbatim}

\begin{figure}[h]
    \centering
    \includegraphics[width=1.0\linewidth]{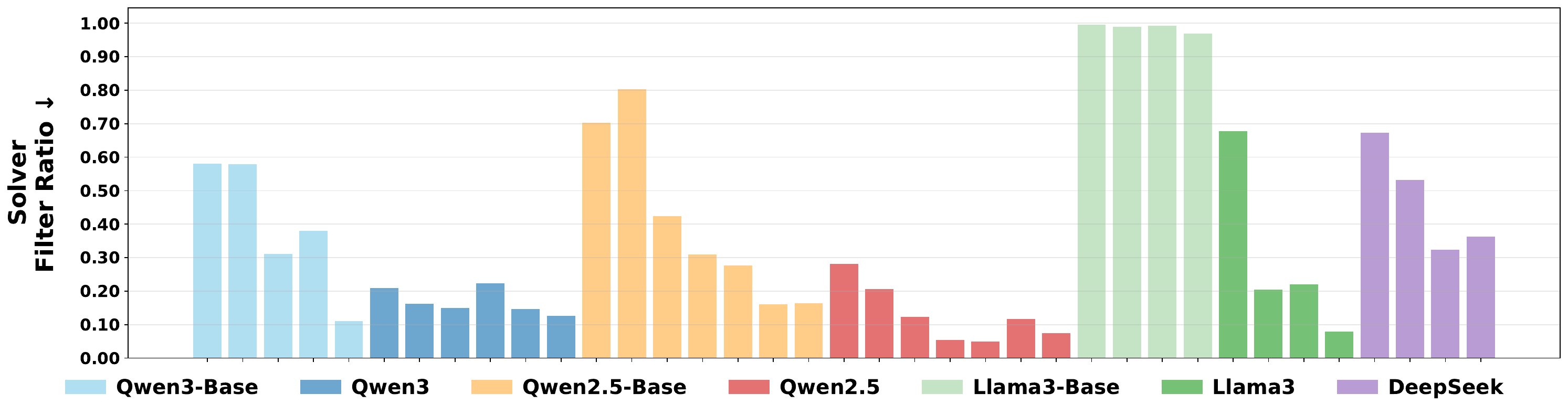}
    \caption{
        \looseness=-10000
        Average ratio of filtered solver outputs for each model over all datasets. Base model families are suffixed by \textbf{-Base}. Models within each family are ordered in increasing size.
    }
    \label{fig:solverbadratio}
\end{figure}

\looseness=-10000
Figure~\ref{fig:solverbadratio} displays the ratio of filtered solver outputs due to not containing a box for answer extraction, averaged across all datasets.

\section{Solver Accuracy by Dataset}\label{appendix:solver-acc}

\begin{figure}
    \centering
    \includegraphics[width=1.0\linewidth]{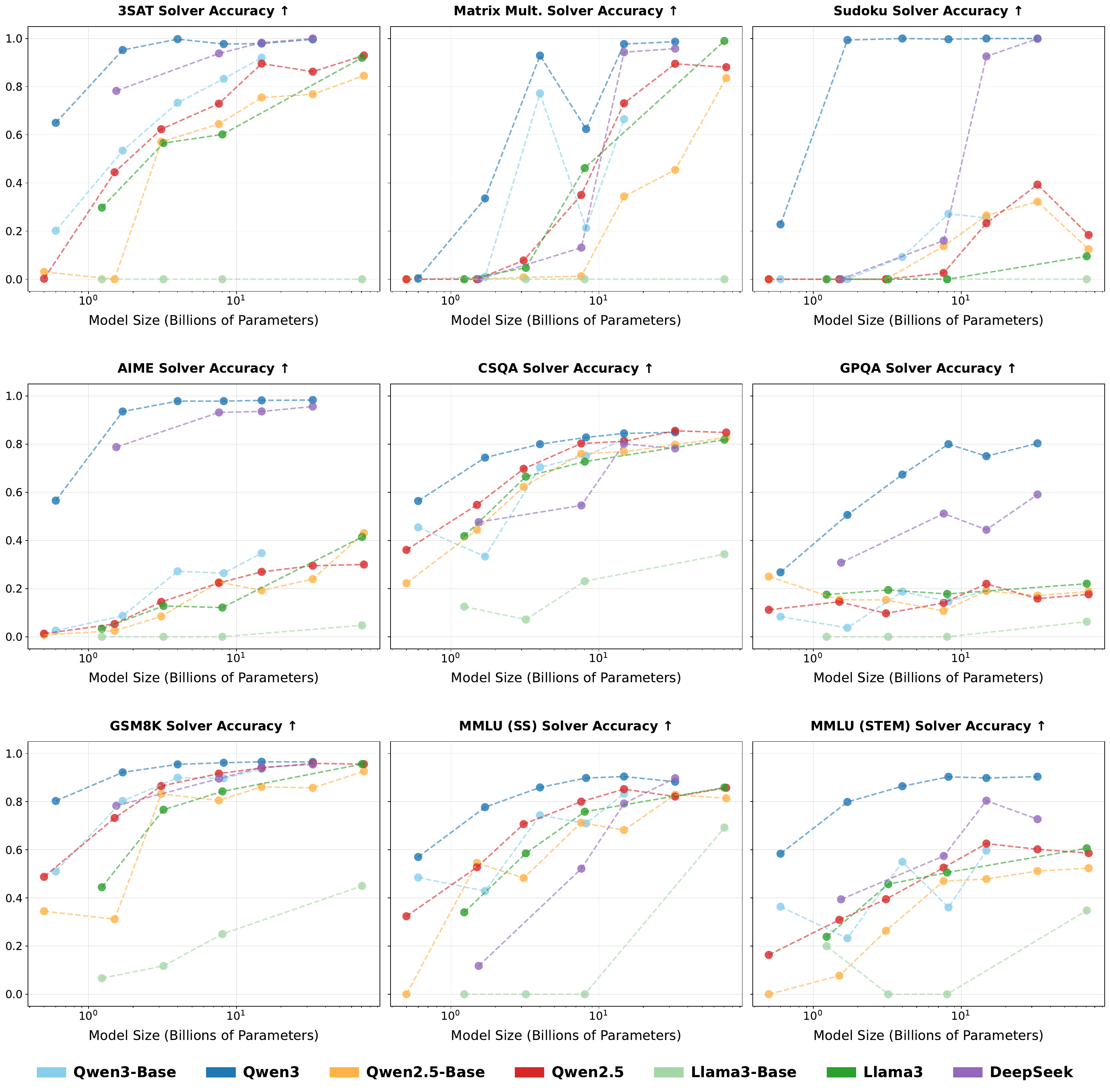}
    \caption{
        \looseness=-10000
        The solver accuracies of 37 models on each dataset.
    }
    \label{fig:solver-per-task}
\end{figure}

\looseness=-10000
Figure~\ref{fig:solver-per-task} shows the solver accuracies of all 37 models on each of our 9 datasets.

\newpage

\section{F1-Score and Precision Visualization}\label{appendix:f1_and_precision}

\begin{figure}
    \centering
    \includegraphics[width=0.925\linewidth]{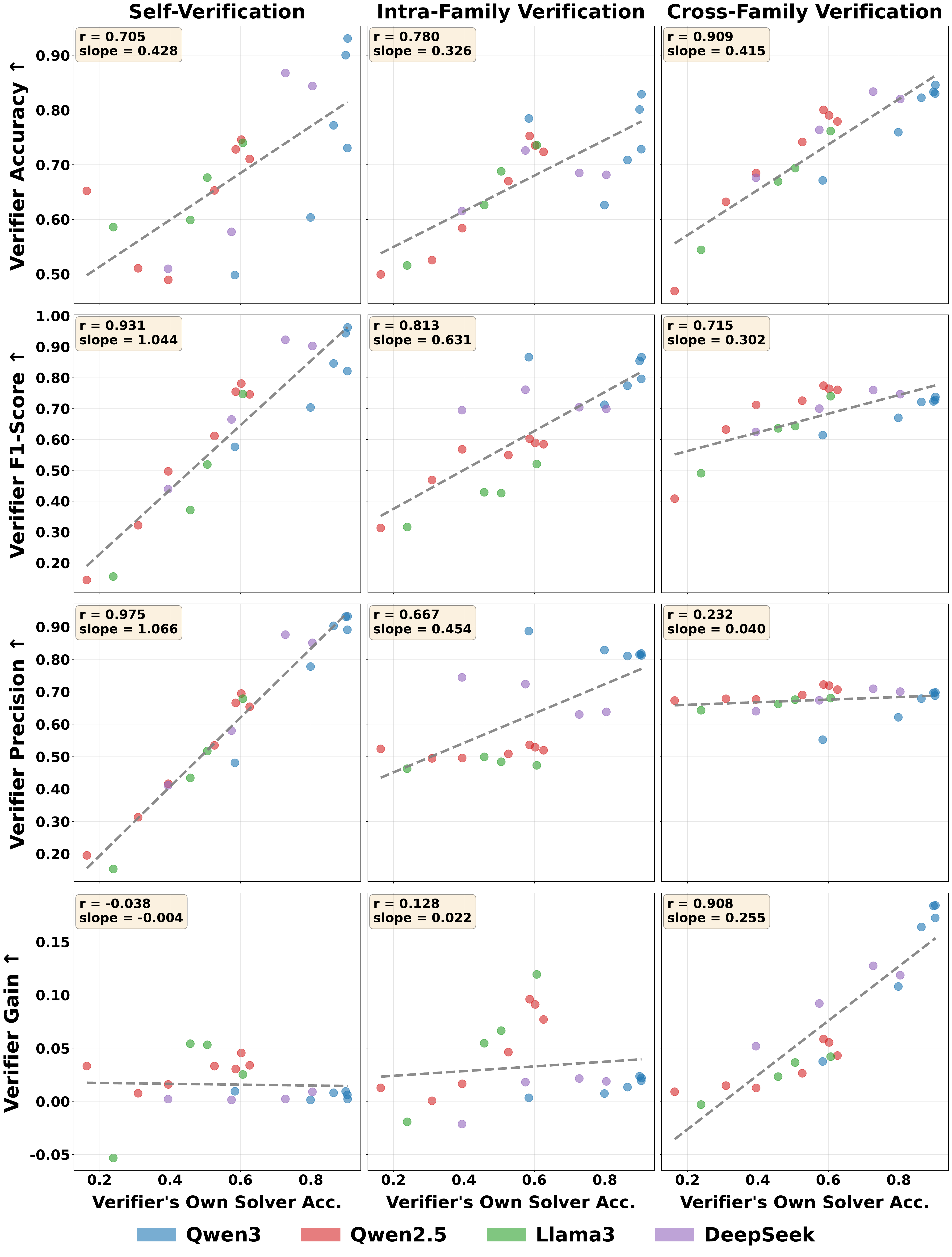}
    \caption{
        \looseness=-10000
        Correlation between each model's verifier metrics (rows) and its own solver accuracy for all 21 post-trained models, averaged over all datasets. Each verifier metric is computed over three settings (columns): self-verification, intra-family verification, and cross-family verification. We use the same set of post-trained models as the set of solver models.
    }
    \label{fig:f1scatter}
\end{figure}

\looseness=-10000
Figure~\ref{fig:cross-dataset-verifier-scatter-solver-acc} shows the correlation between each model's verification ability and its own solver accuracy for all 21 post-trained models. We additionally display verifier F1-Score and precision in Figure~\ref{fig:f1scatter}. 

\looseness=-10000
In comparison to verifier accuracy, while F1-Score also positively correlates with the verifier's own solver accuracy for all verification settings, the slopes decrease from self-verification to intra-family verification, and further decrease for cross-family verification, showing that the increase in false positive rate in Figure~\ref{fig:cross-dataset-verifier-scatter-solver-acc} has a stronger impact on lowering F1-Score than accuracy.

\looseness=-10000
While Section~\ref{sec:bettersolverbetterverifiers} explains the low verifier gains for self- and intra-family verification through close examination of FPR, we additionally plot verifier precision in Figure~\ref{fig:f1scatter}. However, since precision is the expected performance of verifier-based rejection sampling in the limit of infinite sampling and our main metric “verifier gain” is defined in terms of it (Equation~\ref{eqn:verifiergain}), precision does not help explain the differences in verifier gains across verification settings.

\section{Self-Verification Keywords}
\label{appendix:keywords}

To measure the spontaneous self-verification rate of solver outputs (Section~\ref{sec:bettersolverbetterverifiers}), we scan each solver output for the following
case-insensitive keywords (separated by commas):

\begin{quote}
\textit{let me check, let me verify, double-check, let me recalculate, wait, hold on, going back, reconsider, let me re-, mistake, i made, i forgot, i missed,
that doesn't, that's wrong, that's not, this is wrong}
\end{quote}

A solver output is classified as containing spontaneous self-verification if any of these keywords appear in the generated text.

\section{Generation Log-Likelihood as an Alternative Similarity Metric}
\label{appendix:likelihood}

As a complement to the cosine similarity analysis in Section~\ref{sec:similaritycorrelation}, we compute the average per-token log-likelihood that each verifier assigns to each solver's outputs and use this as an alternative similarity metric. A key limitation of this metric is that it conflates two signals: (a)~distributional similarity between the solver and verifier, which is the quantity of interest, and (b)~intrinsic predictability of the solver's text, independent of the evaluating model. For example, short, conventionally structured outputs from \texttt{Llama3} receive higher likelihood from most verifiers than the long reasoning chains with backtracking produced by \texttt{DeepSeek}, regardless of distributional similarity. To isolate~(a), we normalize by solver, subtracting each solver's mean log-likelihood across all verifiers.

\begin{figure}[t]
    \centering
    \includegraphics[width=0.5\linewidth]{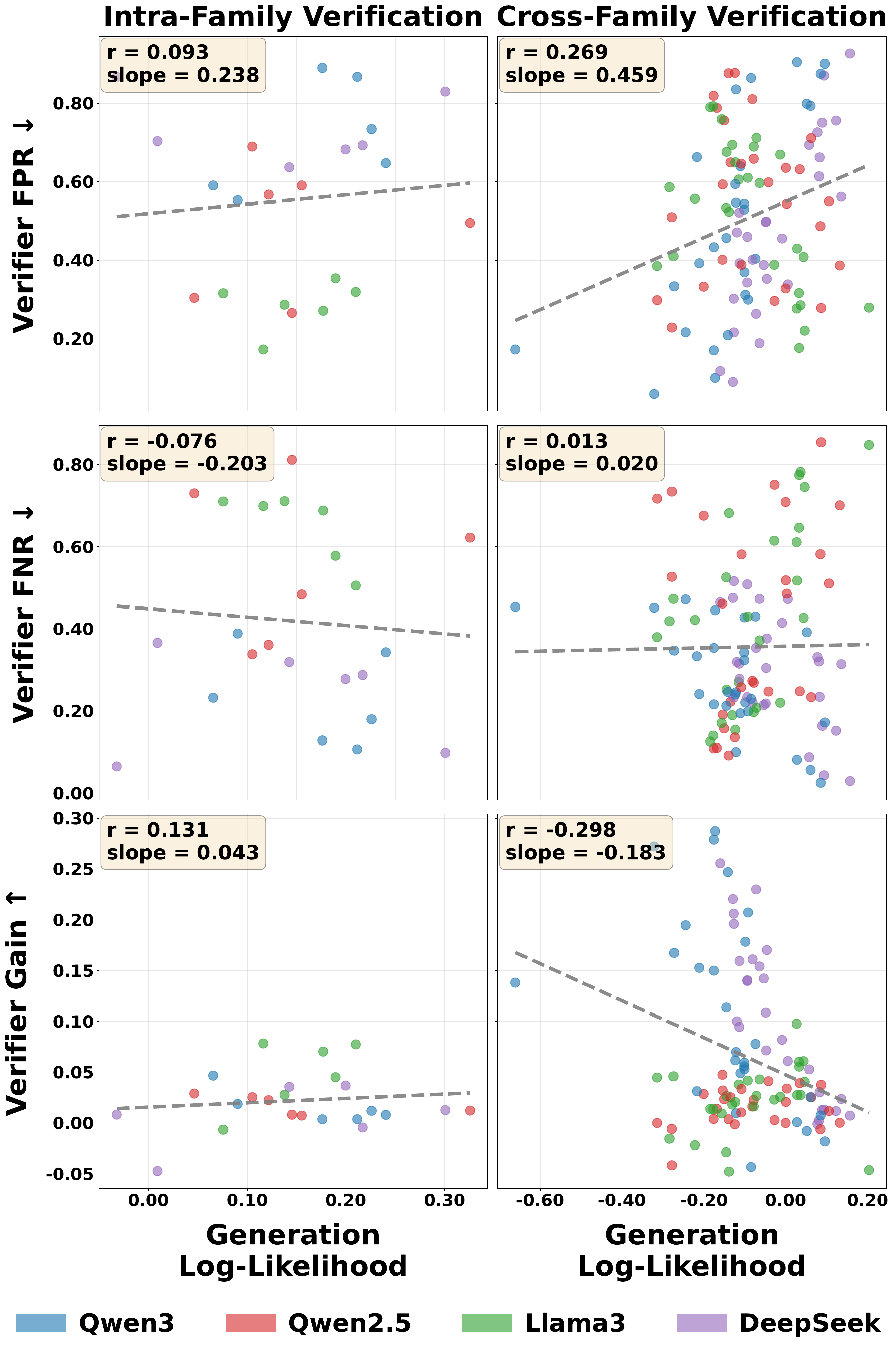}
    \caption{
        \looseness=-10000
        Correlation between verifier metrics and solver-normalized generation log-likelihood for each solver--verifier pair. Each marker is colored based on the
verifier model family. Self-verification is omitted as it yields only one data point per model, too few for reliable correlation analysis.
    }
    \label{fig:similarity-plot-likelihood}
\end{figure}

After normalization, the same directional findings emerge (Figure~\ref{fig:similarity-plot-likelihood}): higher generation log-likelihood correlates with higher FPR (intra-family slope~$= +0.238$, cross-family slope~$= +0.459$) and lower verifier gain (intra-family slope~$= +0.043$, cross-family slope~$= -0.183$), consistent with the cosine similarity results in Figure~\ref{fig:similarity-plot}.

\section{Effect of Reasoning Post-Training on Solver Performance}\label{appendix:posttrainsolver}

\begin{figure}[h]
    \centering
    \includegraphics[width=0.4\linewidth]{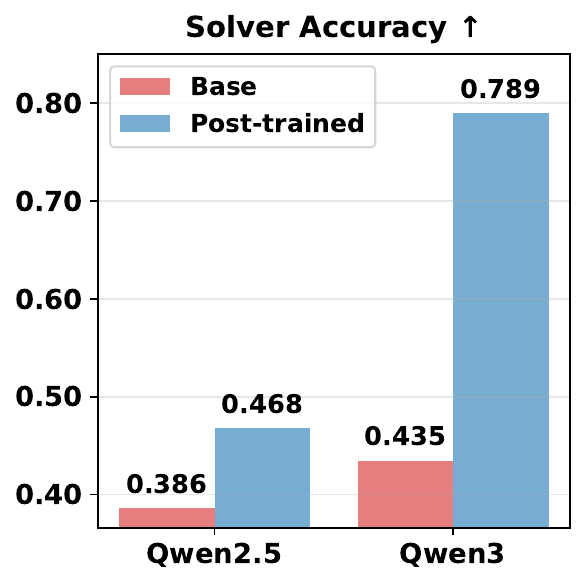}
    \caption{
        \looseness=-10000
        Improvements in solver accuracies of \texttt{Qwen2.5-Base} and \texttt{Qwen3-Base} models from reasoning post-training.
    }
    \label{fig:post-train-solver}
\end{figure}

\looseness=-10000
Figure~\ref{fig:post-train-solver} shows the average improvement in solver accuracies of \texttt{Qwen2.5-Base} and \texttt{Qwen3-Base} families of models from their respective post-training procedures. 

\begin{table}[t]
    \small
    \centering
    \caption{Complete list of each evaluated model's HuggingFace identifier, family, and size.}
    \begin{tabular}{lll}
    \toprule
    \textbf{HuggingFace Identifier} & \textbf{Family} & \textbf{Size} \\
    \midrule
    \color{QwenThreeBase}Qwen/Qwen3-0.6B-Base & \color{QwenThreeBase}Qwen3-Base & \color{QwenThreeBase}0.6B \\
    \color{QwenThreeBase}Qwen/Qwen3-1.7B-Base & \color{QwenThreeBase}Qwen3-Base & \color{QwenThreeBase}1.7B \\
    \color{QwenThreeBase}Qwen/Qwen3-4B-Base & \color{QwenThreeBase}Qwen3-Base & \color{QwenThreeBase}4B \\
    \color{QwenThreeBase}Qwen/Qwen3-8B-Base & \color{QwenThreeBase}Qwen3-Base & \color{QwenThreeBase}8B \\
    \color{QwenThreeBase}Qwen/Qwen3-14B-Base & \color{QwenThreeBase}Qwen3-Base & \color{QwenThreeBase}14B \\
    \color{QwenThree}Qwen/Qwen3-0.6B & \color{QwenThree}Qwen3 & \color{QwenThree}0.6B \\
    \color{QwenThree}Qwen/Qwen3-1.7B & \color{QwenThree}Qwen3 & \color{QwenThree}1.7B \\
    \color{QwenThree}Qwen/Qwen3-4B & \color{QwenThree}Qwen3 & \color{QwenThree}4B \\
    \color{QwenThree}Qwen/Qwen3-8B & \color{QwenThree}Qwen3 & \color{QwenThree}8B \\
    \color{QwenThree}Qwen/Qwen3-14B & \color{QwenThree}Qwen3 & \color{QwenThree}14B \\
    \color{QwenThree}Qwen/Qwen3-32B & \color{QwenThree}Qwen3 & \color{QwenThree}32B \\
    \color{QwenTwoFiveBase}Qwen/Qwen2.5-0.5B & \color{QwenTwoFiveBase}Qwen2.5-Base & \color{QwenTwoFiveBase}0.5B \\
    \color{QwenTwoFiveBase}Qwen/Qwen2.5-1.5B & \color{QwenTwoFiveBase}Qwen2.5-Base & \color{QwenTwoFiveBase}1.5B \\
    \color{QwenTwoFiveBase}Qwen/Qwen2.5-3B & \color{QwenTwoFiveBase}Qwen2.5-Base & \color{QwenTwoFiveBase}3B \\
    \color{QwenTwoFiveBase}Qwen/Qwen2.5-7B & \color{QwenTwoFiveBase}Qwen2.5-Base & \color{QwenTwoFiveBase}7B \\
    \color{QwenTwoFiveBase}Qwen/Qwen2.5-14B & \color{QwenTwoFiveBase}Qwen2.5-Base & \color{QwenTwoFiveBase}14B \\
    \color{QwenTwoFiveBase}Qwen/Qwen2.5-32B & \color{QwenTwoFiveBase}Qwen2.5-Base & \color{QwenTwoFiveBase}32B \\
    \color{QwenTwoFiveBase}Qwen/Qwen2.5-72B & \color{QwenTwoFiveBase}Qwen2.5-Base & \color{QwenTwoFiveBase}72B \\
    \color{QwenTwoFive}Qwen/Qwen2.5-0.5B-Instruct & \color{QwenTwoFive}Qwen2.5 & \color{QwenTwoFive}0.5B \\
    \color{QwenTwoFive}Qwen/Qwen2.5-1.5B-Instruct & \color{QwenTwoFive}Qwen2.5 & \color{QwenTwoFive}1.5B \\
    \color{QwenTwoFive}Qwen/Qwen2.5-3B-Instruct & \color{QwenTwoFive}Qwen2.5 & \color{QwenTwoFive}3B \\
    \color{QwenTwoFive}Qwen/Qwen2.5-7B-Instruct & \color{QwenTwoFive}Qwen2.5 & \color{QwenTwoFive}7B \\
    \color{QwenTwoFive}Qwen/Qwen2.5-14B-Instruct & \color{QwenTwoFive}Qwen2.5 & \color{QwenTwoFive}14B \\
    \color{QwenTwoFive}Qwen/Qwen2.5-32B-Instruct & \color{QwenTwoFive}Qwen2.5 & \color{QwenTwoFive}32B \\
    \color{QwenTwoFive}Qwen/Qwen2.5-72B-Instruct & \color{QwenTwoFive}Qwen2.5 & \color{QwenTwoFive}72B \\
    \color{LlamaThreeBase}meta-llama/Llama-3.2-1B & \color{LlamaThreeBase}Llama3-Base & \color{LlamaThreeBase}1B \\
    \color{LlamaThreeBase}meta-llama/Llama-3.2-3B & \color{LlamaThreeBase}Llama3-Base & \color{LlamaThreeBase}3B \\
    \color{LlamaThreeBase}meta-llama/Llama-3.1-8B & \color{LlamaThreeBase}Llama3-Base & \color{LlamaThreeBase}8B \\
    \color{LlamaThreeBase}meta-llama/Llama-3.1-70B & \color{LlamaThreeBase}Llama3-Base & \color{LlamaThreeBase}70B \\
    \color{LlamaThree}meta-llama/Llama-3.2-1B-Instruct & \color{LlamaThree}Llama3 & \color{LlamaThree}1B \\
    \color{LlamaThree}meta-llama/Llama-3.2-3B-Instruct & \color{LlamaThree}Llama3 & \color{LlamaThree}3B \\
    \color{LlamaThree}meta-llama/Llama-3.1-8B-Instruct & \color{LlamaThree}Llama3 & \color{LlamaThree}8B \\
    \color{LlamaThree}meta-llama/Llama-3.1-70B-Instruct & \color{LlamaThree}Llama3 & \color{LlamaThree}70B \\
    \color{DeepSeek}deepseek-ai/DeepSeek-R1-Distill-Qwen-1.5B & \color{DeepSeek}DeepSeek & \color{DeepSeek}1.5B \\
    \color{DeepSeek}deepseek-ai/DeepSeek-R1-Distill-Qwen-7B & \color{DeepSeek}DeepSeek & \color{DeepSeek}7B \\
    \color{DeepSeek}deepseek-ai/DeepSeek-R1-Distill-Qwen-14B & \color{DeepSeek}DeepSeek & \color{DeepSeek}14B \\
    \color{DeepSeek}deepseek-ai/DeepSeek-R1-Distill-Qwen-32B & \color{DeepSeek}DeepSeek & \color{DeepSeek}32B \\
    \bottomrule
    \end{tabular}
    \label{tab:allmodels}
\end{table}

\begin{table}[h]
    \centering
    \caption{HuggingFace information and sizes of real-world datasets.}
    \begin{tabular}{lllll}
    \toprule
    \textbf{Dataset Name} & \textbf{HuggingFace Identifier} & \textbf{HuggingFace Split} & \textbf{Size} \\
    \midrule
    GSM8K & openai/gsm8k & test & 1319 \\
    AIME & TianHongZXY/aime-1983-2025 & test & 963 \\
    MMLU (STEM) & cais/mmlu & test & 316 \\
    MMLU (Social Sciences) & cais/mmlu & test & 308 \\
    CSQA & tau/commonsense\_qa & validation & 2442 \\
    GPQA & Idavidrein/gpqa & train & 198 \\
    \bottomrule
    \end{tabular}
    \label{tab:alldatasets}
\end{table}

\end{document}